\documentclass[a4paper,fleqn,final]{cas-dc}

\usepackage[T1]{fontenc}
\usepackage[utf8]{inputenc}
\usepackage[main=english,romanian]{babel}
\usepackage{textcomp}
\usepackage{hyperref}
\usepackage{amsmath}
\usepackage{amssymb}
\usepackage{graphicx}
\usepackage{booktabs}
\usepackage{tabularx}
\usepackage{threeparttable}
\usepackage[numbers,sort&compress]{natbib}
\usepackage{longtable}
\usepackage{adjustbox}
\usepackage{caption}
\usepackage{etoolbox}
\usepackage{placeins}
\usepackage{xurl}
\usepackage{microtype}

\makeatletter
\g@addto@macro{\UrlBreaks}{\do\_\do\-\do\/\do\.\do\0\do\1\do\2\do\3\do\4\do\5\do\6\do\7\do\8\do\9}
\makeatother

\renewcommand{\ttfamily}{\rmfamily}
\AtBeginDocument{%
  \urlstyle{same}%
}

\usepackage{xcolor}
\definecolor{linknavy}{RGB}{0,70,127}
\hypersetup{
  colorlinks=true,
  linkcolor=black,      % Fig/Table/footnote markers stay black
  citecolor=linknavy,   % [1], [2], ...
  urlcolor=linknavy,    % DOI, GitHub, mailto
  runcolor=linknavy
}

\AtBeginEnvironment{table}{\let\sffamily\rmfamily}
\AtBeginEnvironment{figure}{\let\sffamily\rmfamily}
\AtBeginEnvironment{table*}{\let\sffamily\rmfamily}
\AtBeginEnvironment{figure*}{\let\sffamily\rmfamily}

\ExplSyntaxOn
\RenewDocumentCommand \firstname {}
	{ \textcolor{black}{\seq_use:Nn \l_stm_au_seq { ~ }} }

\RenewDocumentCommand \emailauthor { m m }
   {
     \int_gincr:N \g_ead_int
     \seq_gput_right:Nn \g_stm_ead_seq
       {
         { \href{mailto:#1}{\rmfamily #1} }
         \parsename { #2 }
         \space(\eadauthor)
       }
     }

\cs_set:Npn \__first_footerline:
{
	\group_begin:
	\small
	\normalfont
	\ifnum\theblind>0\relax
	\else
	\__short_authors: :~
	\fi
	\itshape Preprint~submitted~to~Taylor~\&~Francis
	\group_end:
}

\RenewDocumentCommand \printorcid { } { }

\cs_set:Npn \__first_head:
{
	\parbox[t]{\textwidth}
	{
		\rule{\textwidth}{0pt}
	}
}

\cs_set:Npn \__cas_head:
{
	\parbox{\textwidth}
	{
		\rule{\textwidth}{0pt}
	}
}

\cs_set:Npn \__cas_foot:
{
	\parbox[t]{\textwidth}
	{
	 \rule{\textwidth}{.2pt}\\
	 \small
	 \normalfont
	 \__first_footerline:
	 \hfill Page~\thepage {}~of~ \lastpage
	}
}
\ExplSyntaxOff

\makeatletter
\ps@cas
\makeatother

\begin{document}

% IEEE BST control: show full author names (no ------ dashes)
\makeatletter
\def\bstctlcite{\@ifnextchar[{\@bstctlcite}{\@bstctlcite[@auxout]}}
\def\@bstctlcite[#1]#2{\@bsphack
  \@for\@citeb:=#2\do{%
    \edef\@citeb{\expandafter\@firstofone\@citeb}%
    \if@filesw\immediate\write\csname #1\endcsname{\string\citation{\@citeb}}\fi}%
  \@esphack}
\makeatother
\bstctlcite{IEEEbsTcontrol}

% Short author
\shortauthors{Narimani et al.}
\shorttitle{NAIP Farmland Extent with ResUNet and SAM 3}

% Main title of the paper
\title[mode=title]{Farmland Extent and Visible Boundary Mapping from 1 m NAIP Imagery Using Residual U-Net and Text-Prompted SAM 3 Refinement}

% Authors
\author[1]{Mohammadreza Narimani}
\cormark[1]
\ead{mnarimani@ucdavis.edu}

\author[2]{Vikram Anand}
\ead{vikramranand@gmail.com}

\author[1]{Parastoo Farajpoor}
\ead{pfarajpoor@ucdavis.edu}

\affiliation[1]{organization={Department of Biological and Agricultural Engineering, University of California, Davis},city={Davis},state={CA},postcode={95616},country={USA}}

\affiliation[2]{organization={Del Norte High School},city={San Diego},state={CA},country={USA}}

% Corresponding author
\cortext[1]{Corresponding author}

% Abstract
\begin{abstract}
Agricultural fields provide the spatial framework through which crop condition, production, management, and land conversion are interpreted, yet current field maps are often proprietary, incomplete, or outdated. This study develops a reproducible workflow for mapping farmland extent and its visible boundaries from 1\,m National Agriculture Imagery Program (NAIP) RGB imagery. Thirty-seven scenes were selected to represent open cropland, peri-urban interfaces, semi-arid irrigation geometries, and fragmented agricultural mosaics. Farmland polygons were digitized in the Computer Vision Annotation Tool (CVAT), converted to binary masks, and paired with non-overlapping 256\,$\times$\,256 pixel image patches. The resulting 5{,}698 samples were partitioned by source scene into 3{,}850 training, 770 validation, and 1{,}078 test patches, thereby preventing patches from the same scene from crossing data partitions. A deep residual U-Net (ResUNet) was trained with a Dice-dominant objective, $\mathcal{L}=2.5(1-\mathrm{Dice})+\mathrm{BCE}$, to reduce the influence of background-dominated urban--agricultural patches. On the held-out test set, the model achieved an accuracy of 0.8808, intersection over union of 0.8605, Dice coefficient of 0.9234, precision of 0.8766, and recall of 0.9794. The combination of high overlap and high recall indicates that most labeled farmland was retained, while remaining errors were concentrated along visually ambiguous roads, exposed soil, and developed edges. A frozen Segment Anything Model 3 (SAM 3) branch was then applied to each patch with the text prompt ``agricultural farmland field.'' SAM 3 concept masks were unioned and fused with the thresholded ResUNet output by logical OR. On selected difficult examples, the fused prediction improved Dice from 0.858 to 0.955 for orchard-row structure and from 0.804 to 0.903 for fragmented parcels. Sliding-window inference and stitching produced coherent regional masks in contrasting landscapes, with example tile-level Dice scores of 0.898 and 0.919. The results show that a compact domain-trained network can provide stable farmland segmentation while a frozen, text-prompted foundation model supplies complementary concept masks for difficult local structures. The approach is intended as a semantic farmland-extent product rather than a substitute for legal cadastral boundaries, and it provides a practical basis for crop-area accounting, agricultural monitoring, and analysis of farmland conversion where current parcel layers are unavailable.
\end{abstract}

% Keywords

\maketitle

% Main text

\section{Introduction}\label{sec:introduction}

Remote sensing in agriculture is often described as a progression in spatial scale, but it is more accurately understood as a progression in the decisions that become possible. At the leaf scale, radiative-transfer models established that measured reflectance can be linked to biochemical and structural properties \citep{jacquemoud1990prospect}. Recent grapevine studies have extended that relationship with data-driven multi-trait models that estimate nutrients and physiological attributes from hyperspectral observations \citep{farajpoor2025multitrait}, while attention-based forward models have shown that measured traits can also be used to reconstruct leaf reflectance spectra \citep{farajpoor2026attention}. In plant-health monitoring, imaging spectroscopy and machine learning have moved disease and stress detection toward earlier, pre-visual diagnosis \citep{sankaran2010review,mahlein2016plant}. For example, leaf-level spectra have been used to identify branched broomrape stress in tomato before the signal becomes readily visible at canopy scale \citep{narimani2025leaf}.

The same logic extends upward. Unmanned aerial vehicles translate plant-level signals into spatial patterns across rows and management zones, offering a useful compromise between detail, coverage, and revisit frequency \citep{maes2019uav}. Drone-based multispectral time series have consequently been used to detect parasitic-weed stress in tomato fields using deep sequence models \citep{narimani2024drone}. Satellite observations broaden the domain again: they support crop monitoring, yield estimation, phenology, and regional assessment over large areas \citep{weiss2020metareview,segarra2020sentinel}. Recent work has used Sentinel-2 time series for crop-stress detection \citep{narimani2025satellite}, synthesized the rapidly expanding literature on Sentinel-2 yield estimation \citep{narimani2026sentinelreview}, and explored analysis-ready geospatial embeddings for field-scale tomato-system mapping \citep{narimani2026alphaearth}. Across these scales, the sensor changes, but a common analytical question remains: what is the spatial unit to which a trait, stress signal, crop label, or management decision should be assigned?

For most operational agricultural applications, that unit is the field. Field geometry provides the support over which pixel observations are summarized, temporal profiles are compared, crop areas are calculated, and management or policy actions are targeted. Reliable field maps are therefore prerequisites for precision-agriculture services, crop statistics, land-use accounting, carbon and water assessment, insurance, and monitoring the conversion of farmland to residential or industrial uses \citep{mulla2013twentyfive,sishodia2020applications}. Yet field boundaries are not universally equivalent to cadastral parcels, and they are not always maintained as current, accessible digital data. Ownership boundaries may be invisible in imagery; conversely, visible management boundaries may change within a legal parcel. Manual digitization can resolve these distinctions locally but is expensive and difficult to update. Large benchmark and mapping efforts have underscored both the demand for field-level products and the cost of producing them at scale \citep{kerner2025fields,robinson2026global}.

High-resolution aerial photography offers a complementary route to medium-resolution satellite mapping. The U.S. Department of Agriculture's National Agriculture Imagery Program (NAIP) provides broad, repeatedly acquired aerial coverage of the United States \citep{usda2026naip}. The imagery used in this study has a 1\,m ground sampling distance, fine enough to resolve narrow roads, field margins, orchard texture, buildings, and many within-scene transitions that are mixed within a 10\,m satellite pixel. That detail is valuable, but it also makes segmentation harder. A model must separate farmland from visually similar bare ground, lawns, roadside vegetation, industrial lots, and construction areas; preserve long narrow gaps between adjacent parcels; and remain stable across crop color, soil exposure, irrigation geometry, and peri-urban development. These are not merely spectral classification problems. They require contextual and boundary reasoning under strong class imbalance.

Convolutional encoder--decoder networks provide a practical foundation for this task. U-Net introduced a compact architecture in which a contracting path learns context and skip connections restore spatial detail \citep{ronneberger2015unet}. Residual learning improved optimization of deeper networks \citep{he2016resnet}, and residual U-Net variants combined both ideas for dense prediction \citep{zhang2018resunet}. In agricultural boundary mapping, fully convolutional, residual, recurrent, multi-task, and super-resolution approaches have progressively improved field-extent and edge detection \citep{masoud2020delineation,waldner2020deepedge,taravat2021advanced,zhang2021recurrent}. Nevertheless, domain-trained convolutional networks can under-segment low-contrast fields, merge adjacent parcels, or smooth small structures when labels are limited or landscapes shift.

Promptable foundation models offer a different source of prior knowledge. The original Segment Anything Model (SAM) learned class-agnostic segmentation from a large mask corpus \citep{kirillov2023sam}, SAM 2 extended promptable segmentation to images and video \citep{ravi2024sam2}, and SAM 3 added concept prompts that can retrieve all instances matching a short noun phrase \citep{carion2025sam3}. Remote-sensing studies have shown both the potential and the limitations of direct SAM transfer: generic visual priors can improve mask generation, but scale, spectral appearance, dense object packing, and ambiguous boundaries often require domain adaptation or task-specific guidance \citep{osco2023samrs,ao2024sempnet}. Recent smallholder experiments further show that zero-shot SAM performance depends strongly on image scale, temporal information, and field size \citep{tripathy2026zeroshot}.

This study investigates a deliberately simple hybrid: a domain-trained ResUNet supplies a stable farmland probability map, and a frozen, text-prompted SAM 3 branch supplies additional concept masks. Their outputs are fused by logical union and stitched into regional binary maps. The target is farmland extent, with visible boundaries represented implicitly by transitions between farmland and background; the method does not claim to recover legal cadastral lines or to separate every adjacent field as an individual polygon instance.

The study is organized around three research questions:
\begin{enumerate}
\item How accurately does a ResUNet segment farmland extent from 1\,m NAIP RGB imagery when training, validation, and test partitions are separated at the scene level?
\item On selected difficult patches, can text-prompted SAM 3 recover farmland omitted by ResUNet without retraining the foundation model?
\item Can patch predictions be stitched into coherent regional masks across regular field grids and complex peri-urban scenes?
\end{enumerate}

The study contributes: (i) a scene-diverse NAIP annotation and patch-construction protocol designed around landscape heterogeneity; (ii) a residual encoder--decoder trained with a Dice-dominant objective for background-heavy scenes; (iii) a transparent text-prompted SAM 3 refinement rule that can be interpreted directly at the mask level; and (iv) regional demonstrations that connect patch-scale evaluation to stitched mapping across regular and peri-urban agricultural landscapes. The contribution is methodological and empirical: it examines a simple, auditable hybrid rather than claiming universal superiority over alternative segmentation systems.

\section{Related work}\label{sec:related}

\subsection{Agricultural remote sensing from leaves to regional products}

Agricultural remote sensing has developed along two coupled axes: richer measurement of plant function and broader spatial coverage. At proximal scales, hyperspectral measurements can retrieve or predict pigments, water-related properties, nutrient status, and stress responses because biochemical absorption and scattering shape the spectral curve \citep{jacquemoud1990prospect,farajpoor2025multitrait}. Disease-detection reviews have repeatedly emphasized, however, that laboratory separability does not automatically transfer to field conditions, where illumination, canopy geometry, background, phenology, and concurrent stressors alter the signal \citep{sankaran2010review,mahlein2016plant}. The branched-broomrape studies cited above illustrate this scale transition: leaf spectroscopy can reveal early host response \citep{narimani2025leaf}, UAV multispectral imagery can map canopy-level temporal patterns \citep{narimani2024drone}, and Sentinel-2 time series can extend detection to multiple fields \citep{narimani2025satellite}.

The expansion from proximal sensing to Earth observation also changes the role of spatial support. Leaf observations are tied to individual samples; drone observations are often summarized by row or plot; satellite data are commonly aggregated by field. Reviews of precision agriculture and deep learning show that the value of remote sensing depends not only on predictive accuracy but also on whether products align with management units and can be transferred across seasons and locations \citep{mulla2013twentyfive,kamilaris2018deep,weiss2020metareview}. Sentinel-2's spatial, temporal, and spectral design has made it central to crop mapping and yield studies \citep{segarra2020sentinel,narimani2026sentinelreview}, while geospatial embeddings seek to package multi-source, multi-temporal context into reusable representations \citep{narimani2026alphaearth}. In all cases, a missing or outdated field layer limits downstream analysis: pixels cannot be reliably aggregated, neighboring crops may be mixed, and change cannot be attributed to a stable unit.

\subsection{Field-extent and boundary extraction}

Early automated delineation approaches used edge filters, watershed segmentation, region growing, or object-based image analysis. Multi-temporal imagery can improve these methods because crop development creates temporal contrast between adjacent fields even when a single date is ambiguous \citep{watkins2019automating}. Conventional methods remain useful where edges are strong and landscape rules are stable, but they often require local thresholds and post-processing, and they can confuse roads, tree lines, irrigation features, and within-field texture with parcel boundaries.

Deep learning reframed the task as dense prediction. Persello and colleagues combined fully convolutional field detection with combinatorial grouping in smallholder landscapes \citep{persello2019delineation}, while Masoud, Persello, and Tolpekin used super-resolution contour refinement to improve boundaries from Sentinel-2 imagery \citep{masoud2020delineation}. Waldner and Diakogiannis predicted field extent, boundary probability, and distance-to-boundary jointly, demonstrating that correlated tasks and multi-date consensus can improve transfer across sensors, resolutions, locations, and years \citep{waldner2020deepedge}. Their later DECODE framework consolidated seasonal predictions and converted them to field instances with hierarchical watershed segmentation \citep{waldner2021decode}. Other work compared U-Net and ResU-Net variants over large agricultural areas \citep{taravat2021advanced}, incorporated recurrent residual units \citep{zhang2021recurrent}, and used transfer learning and weak supervision to address smallholder landscapes where labels are scarce \citep{wang2022unlocking}.

Two developments are especially relevant to the present study. First, open datasets are making experimental design more transparent. AI4Boundaries harmonized field labels with Sentinel-2 and aerial photography across several European countries \citep{dandrimont2023ai4boundaries}, and Fields of the World broadened benchmark coverage and emphasized global variation in field size and morphology \citep{kerner2025fields}. Second, the literature increasingly distinguishes extent, edge, and instance outputs. A binary field mask can achieve high area overlap while still merging adjacent parcels; an edge map can locate boundaries but fail to close polygons; an instance product requires grouping and topology. Reviews of farmland boundary extraction stress that metrics must match the output type and that results from different definitions are not directly comparable \citep{wang2023survey}. This distinction motivates the terminology used here: our model produces a binary farmland-extent mask, and ``boundary segmentation'' refers to the visible foreground--background transition, not a cadastral or instance-level product.

\subsection{Residual encoder--decoders and promptable foundation models}

U-Net's skip-connected encoder--decoder remains widely used because it balances context and localization with modest computational demands \citep{ronneberger2015unet}. Residual blocks improve gradient propagation and permit deeper feature extraction \citep{he2016resnet}, while the deep residual U-Net used for road extraction provides a direct architectural precedent for preserving thin, spatially extended targets in overhead imagery \citep{zhang2018resunet}. DeepLabV3+ offers another strong segmentation family through atrous spatial pyramid pooling and decoder refinement \citep{chen2018deeplab}; it is therefore an important future baseline for this project. For imbalanced segmentation, overlap-based objectives such as generalized Dice reduce the dominance of background pixels \citep{sudre2017dice}.

Foundation models change the balance between domain specificity and broad visual priors. SAM generates masks from geometric prompts without semantic class labels \citep{kirillov2023sam}. Remote-sensing adaptations have used prompt engineering, parameter-efficient tuning, auxiliary classifiers, and feature fusion to compensate for domain shift \citep{osco2023samrs,ao2024sempnet}. SAM 3 is particularly relevant because concept prompting permits a short text phrase to retrieve candidate masks \citep{carion2025sam3}. That capability removes the need to generate point or box prompts manually, but it does not guarantee agricultural correctness: ``field'' can refer to grass, recreational areas, or open land, and a concept mask can cross management boundaries that are weakly visible.

Farmland-specific evidence helps define the gap. Fully convolutional field masks can be converted into more coherent field objects through grouping and post-processing \citep{persello2019delineation}, while zero-shot SAM experiments in very small fields show that performance depends strongly on image scale and temporal fusion \citep{tripathy2026zeroshot}. The present design keeps SAM 3 frozen and uses it as a second-stage candidate generator, while the task-trained ResUNet remains the primary predictor. This division of labor tests whether open-vocabulary concept masks can complement domain learning through a transparent fusion rule without introducing another farmland-specific training stage.

\section{Materials and methods}\label{sec:methods}

\subsection{Problem definition and workflow}

Let $I \in [0,1]^{H \times W \times 3}$ denote a normalized NAIP RGB image and $Y \in \{0,1\}^{H \times W}$ the reference mask, where 1 represents annotated farmland and 0 represents background. The model estimates a probability surface $P_{r} = f_{\theta}(I)$ and a binary mask $M_{r}$. The target is semantic farmland extent. Boundaries are implied by the spatial transition between 1 and 0; no instance identifier is assigned to individual parcels.

\begin{figure*}[t]
\centering
\includegraphics[width=\textwidth]{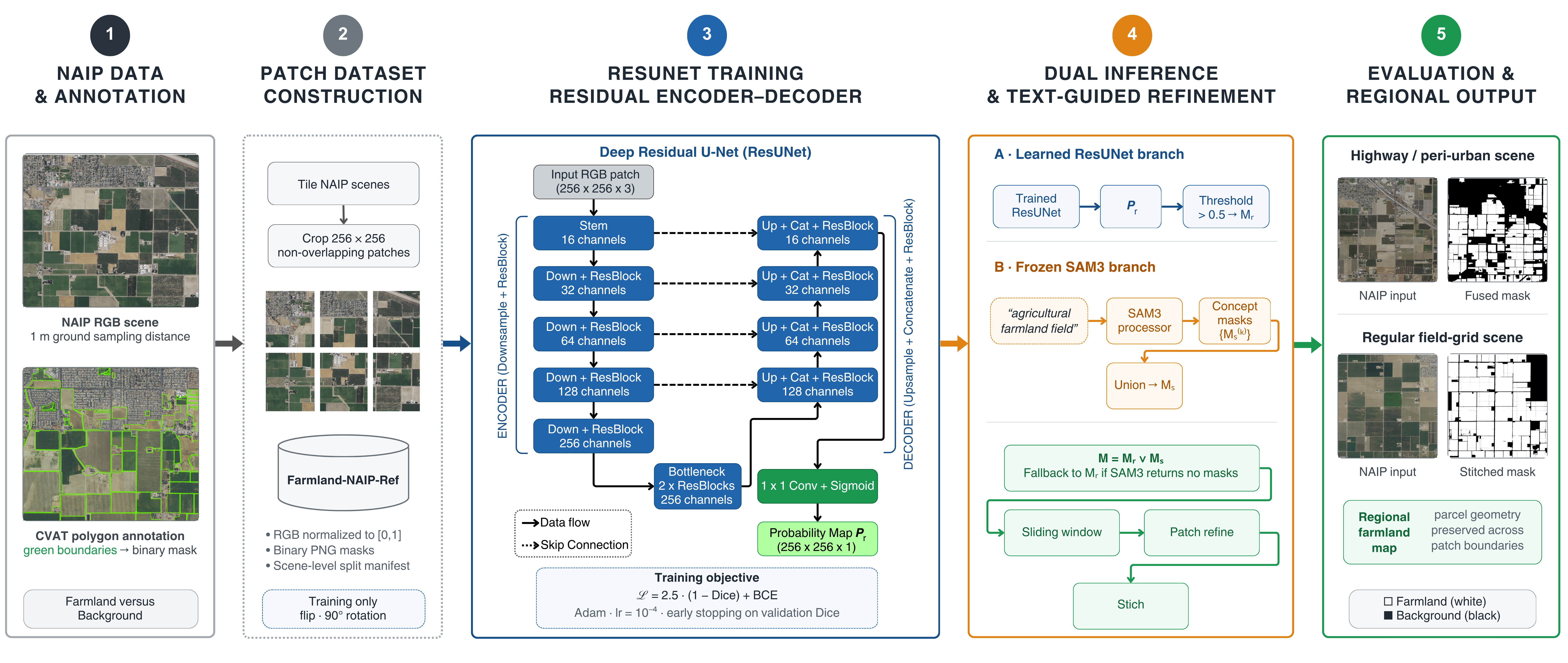}
\caption{Hybrid ResUNet and SAM 3 workflow for farmland-extent and visible-boundary segmentation from NAIP imagery. The learned branch predicts $P_{r}$ and thresholds it to $M_{r}$; the frozen SAM 3 branch returns concept masks for the prompt ``agricultural farmland field.'' Masks are unioned, fused by logical OR, and stitched at scene scale.}
\label{fig:workflow}
\end{figure*}

Figure~\ref{fig:workflow} summarizes five stages: NAIP acquisition and polygon annotation; patch construction; ResUNet training; dual inference with frozen SAM 3 and logical-OR fusion; and sliding-window stitching into regional masks.

\subsection{NAIP imagery and landscape sampling}

The imagery consists of NAIP aerial photographs at 1\,m ground sampling distance. Although the source products include RGB and, where available, near-infrared information, only the three visible bands were used. Restricting the model to RGB simplifies deployment on standard imagery and aligns with the three-channel input expected by common pretrained vision pipelines, but it also removes vegetation information that could help distinguish crops from bare soil or non-crop vegetation.

Thirty-seven source scenes were selected across four landscape contexts: (i) open farmland dominated by large parcels; (ii) peri-urban areas where crop fields meet roads, warehouses, schools, and housing; (iii) semi-arid scenes whose geometry is shaped by irrigation and exposed soil; and (iv) fragmented agricultural mosaics with smaller, irregular fields. The sampling was purposive rather than statistically representative. Its role was to expose the model to different boundary and background conditions, not to support claims about performance across all U.S. regions or acquisition years.

\subsection{Reference-mask construction}

Farmland was manually delineated in the Computer Vision Annotation Tool (CVAT) using polygon tools without a vertex cap. Annotators traced visible crop-area boundaries rather than legal property lines. Roads, buildings, water, residential land, industrial surfaces, and other non-crop areas were assigned to background. Polygon exports were rasterized as binary PNG masks at the image grid: white for farmland and black for background.

\begin{figure*}[t]
\centering
\includegraphics[width=\textwidth]{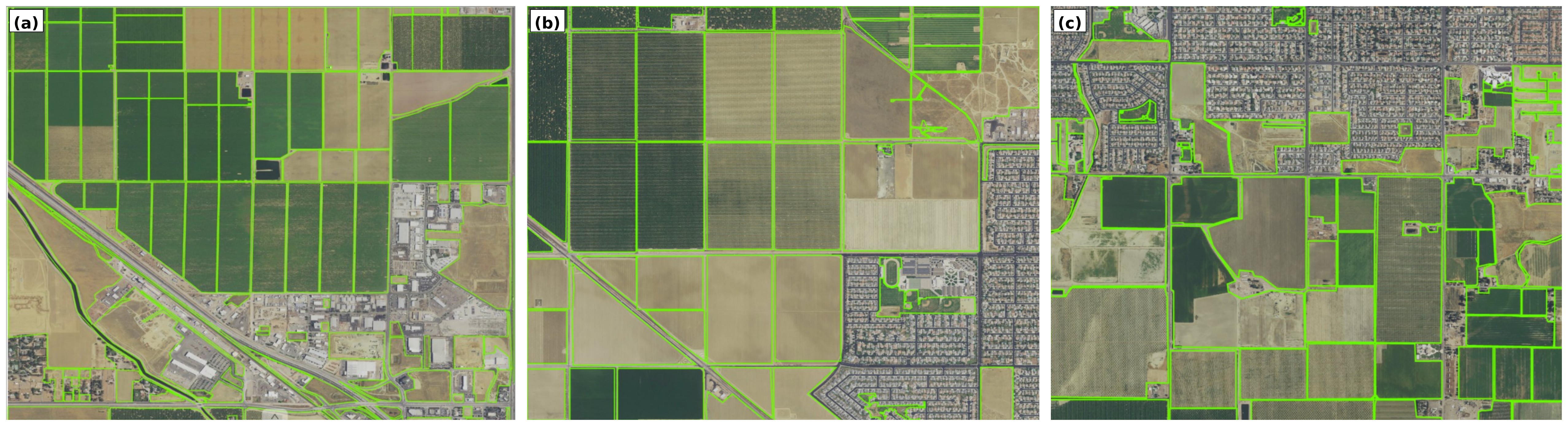}
\caption{Representative CVAT polygon annotations. (a) Highway and industrial edge. (b) Residential and school interface. (c) Urban--farmland mosaic. Green outlines indicate manually traced farmland polygons.}
\label{fig:annotations}
\end{figure*}

Figure~\ref{fig:annotations} illustrates three annotation contexts. The highway/industrial example contains long transportation edges and developed surfaces; the residential example tests separation of fields from housing and a school complex; and the urban--farmland example combines fragmented fields with dense built-up texture. These scenes were included because easy, agriculture-dominant tiles alone would underrepresent the false-positive conditions expected in operational mapping.

\subsection{Patch generation and scene-level partitioning}

Each image--mask pair was cropped into non-overlapping 256\,$\times$\,256 pixel patches. At 1\,m ground sampling distance, a patch represents approximately 256\,$\times$\,256\,m, providing local context for roads, field edges, and portions of large parcels while remaining small enough for batch training on a Colab NVIDIA T4 GPU. RGB values were normalized to $[0,1]$, and masks were loaded as one-channel binary arrays.

The 5{,}698 patches were partitioned by source scene rather than randomly by patch. All patches from a scene were assigned to only one split, reducing the risk that nearly adjacent image content would appear in both training and evaluation. The resulting split contained 3{,}850 training, 770 validation, and 1{,}078 test patches. During training, random horizontal and vertical flips and rotations in 90\textdegree{} increments were applied. Table~\ref{tab:design} summarizes the experimental design.

\begin{table*}[t]
\centering
\caption{Experimental design and reproducibility-critical settings.}
\label{tab:design}
\begin{tabular*}{\textwidth}{@{\extracolsep{\fill}}ll@{}}
\toprule
\textbf{Component} & \textbf{Setting} \\
\midrule
Source imagery & NAIP RGB, 1\,m ground sampling distance; NIR not used \\
Reference classes & Farmland (1) and background (0) \\
Source scenes & 37 scenes across open, peri-urban, semi-arid/irrigated, and fragmented contexts \\
Patch construction & 256\,$\times$\,256 pixels; non-overlapping \\
Partitions & Train 3{,}850; validation 770; test 1{,}078; scene-level separation \\
Training augmentation & Horizontal/vertical flips and 90\textdegree{} rotations \\
Framework and hardware & TensorFlow/Keras; Google Colab NVIDIA T4 \\
Optimization & Adam; learning rate $10^{-4}$; batch size 8; seed 2019; up to 100 epochs \\
Model selection & Validation-Dice checkpointing, learning-rate reduction, and early stopping \\
ResUNet threshold & $P_{r} > 0.5$ \\
SAM 3 prompt & ``agricultural farmland field'' \\
Fusion & $M = M_{r} \vee M_{s}$; fallback to $M_{r}$ when SAM 3 returns no mask \\
\bottomrule
\end{tabular*}
\end{table*}

\subsection{Residual U-Net architecture}

The ResUNet follows a symmetric residual encoder--decoder. A 16-channel stem applies convolution, batch normalization, rectified linear activation, and a projected shortcut. Four downsampling stages increase channel depth to 32, 64, 128, and 256. The bottleneck contains two residual blocks at 256 channels. In the decoder, nearest-neighbor upsampling is followed by concatenation with the corresponding encoder feature map and residual processing at 128, 64, 32, and 16 channels. A $1 \times 1$ convolution with sigmoid activation produces $P_{r} \in [0,1]^{256 \times 256}$.

Residual shortcuts address optimization degradation in deeper networks \citep{he2016resnet}, whereas long U-Net skip connections return high-resolution structure to the decoder \citep{ronneberger2015unet}. This combination is well suited to overhead imagery, where the network must use broad contextual cues to identify agricultural land while retaining narrow exclusions such as roads and field margins \citep{zhang2018resunet}.

\subsection{Loss function and training}

For a batch of reference values $y_{i}$ and predicted probabilities $\widehat{y}_{i}$, the smoothed Dice coefficient was implemented as
\begin{equation}
\mathrm{Dice}\left(Y,\widehat{Y}\right) = \frac{2\sum_{i} y_{i}\widehat{y}_{i} + 1}{\sum_{i} y_{i} + \sum_{i} \widehat{y}_{i} + 1}
\end{equation}
The training objective combined Dice loss with binary cross-entropy (BCE):
\begin{equation}
\mathcal{L} = 2.5\left(1 - \mathrm{Dice}\left(Y,\widehat{Y}\right)\right) + \mathrm{BCE}\left(Y,\widehat{Y}\right)
\end{equation}
The 2.5 multiplier makes overlap error the dominant term. This choice was motivated by patches in which background occupies most pixels; under such imbalance, a pixelwise loss can reward conservative predictions that omit large farmland regions. Dice-based losses are widely used to counter this behavior because they normalize overlap by foreground size \citep{sudre2017dice}. BCE was retained to provide stable pixelwise gradients and probability calibration.

Training used Adam with an initial learning rate of $10^{-4}$, batch size 8, and random seed 2019 for up to 100 epochs. The best checkpoint was selected by validation Dice. ReduceLROnPlateau and EarlyStopping callbacks controlled learning-rate decay and termination. ModelCheckpoint retained the weights associated with the highest validation Dice, while ReduceLROnPlateau and EarlyStopping controlled learning-rate decay and termination when validation improvement slowed.

\subsection{Text-prompted SAM 3 refinement}

SAM 3 was loaded as a frozen model through the Sam3\-Processor interface on a CUDA backend. For each RGB patch, the ResUNet probability map was thresholded at 0.5:
\begin{equation}
M_{r} = \mathbb{1}\left[P_{r} > 0.5\right]
\end{equation}
The same patch was passed to SAM 3 with the exact text prompt ``agricultural farmland field.'' If the processor returned $K$ masks, they were combined by pixelwise union:
\begin{equation}
M_{s} = \bigvee_{k=1}^{K} M_{s}^{(k)}
\end{equation}
The final fused mask was
\begin{equation}
M = M_{r} \vee M_{s}
\end{equation}
If SAM 3 returned no masks or raised an inference exception, $M_{r}$ was returned unchanged. Logical OR was chosen as a recall-preserving prototype: it cannot delete a positive ResUNet pixel and can only add areas proposed by SAM 3. This property makes the operation easy to interpret, but it also means that SAM 3 false positives cannot be corrected by fusion. The rule should therefore be viewed as an initial design to be tested against confidence weighting, intersection gating, boundary-aware fusion, and selective invocation on uncertain patches.

\subsection{Tile-scale inference and stitching}

Full scenes were processed as grids of 256\,$\times$\,256 windows. Each window was normalized, segmented by ResUNet, refined by SAM 3, and placed at its source coordinate in a scene-sized raster. Because the study uses binary semantic masks, stitching consisted of assembling window outputs on the original grid rather than resolving overlapping instance polygons. The selected examples indicate whether patchwise predictions remain spatially coherent across long field edges and mixed urban--agricultural regions. Runtime and memory were not systematically recorded and remain necessary for assessing operational feasibility.

\subsection{Evaluation metrics}

Performance was summarized with pixel accuracy, intersection over union (IoU), Dice coefficient, precision, recall, and the combined training loss. For binary predictions with true positives $TP$, false positives $FP$, and false negatives $FN$, the principal overlap metrics were
\begin{equation}
\mathrm{IoU} = \frac{TP}{TP + FP + FN}
\end{equation}
and
\begin{equation}
\mathrm{Dice} = \frac{2\,TP}{2\,TP + FP + FN}
\end{equation}
Accuracy was retained for completeness, but Dice and IoU were treated as the primary measures because they directly quantify foreground overlap and are less dominated by the background class. Precision and recall were interpreted together to distinguish commission errors from farmland omission. Aggregate metrics were calculated for the training, validation, and test partitions, and batch-level Dice distributions were examined to characterize score variability. The stitched regional examples were evaluated against their corresponding binary reference masks using the same overlap definition.

\section{Results}\label{sec:results}

\subsection{Patch-scale ResUNet performance}

Table~\ref{tab:performance} summarizes aggregate ResUNet performance. Validation and test Dice were both 0.9234, while their IoU values differed by less than 0.001, indicating closely matched behavior across the two scene-level hold-outs. On the test partition, accuracy was 0.8808, IoU was 0.8605, and Dice was 0.9234. Recall reached 0.9794, compared with precision of 0.8766. The model therefore omitted relatively little annotated farmland but included some visually similar non-farmland pixels, especially around developed margins and linear infrastructure. This asymmetry is consistent with the Dice-dominant objective, which assigns a comparatively high cost to missed farmland.

\begin{table*}[t]
\centering
\caption{ResUNet performance by data partition.}
\label{tab:performance}
\begin{tabular*}{\textwidth}{@{\extracolsep{\fill}}lrrrrrr@{}}
\toprule
\textbf{Split} & \textbf{Accuracy} & \textbf{IoU} & \textbf{Dice} & \textbf{Precision} & \textbf{Recall} & \textbf{Loss} \\
\midrule
Train      & 0.9052 & 0.8895 & 0.9404 & 0.9154 & 0.9689 & 0.0596 \\
Validation & 0.8825 & 0.8613 & 0.9234 & 0.8800 & 0.9757 & 0.0766 \\
Test       & 0.8808 & 0.8605 & 0.9234 & 0.8766 & 0.9794 & 0.0766 \\
\bottomrule
\end{tabular*}
\end{table*}

The train-to-test Dice difference was 0.0170. Together with the similar validation and test results, this pattern suggests that scene-level separation limited optimistic leakage while preserving stable performance within the sampled collection. The result should be interpreted at the level of the four represented landscape contexts rather than as a claim of uniform performance across all NAIP acquisitions.

\subsection{Distribution of overlap scores}

\begin{figure*}[t]
\centering
\includegraphics[width=\textwidth]{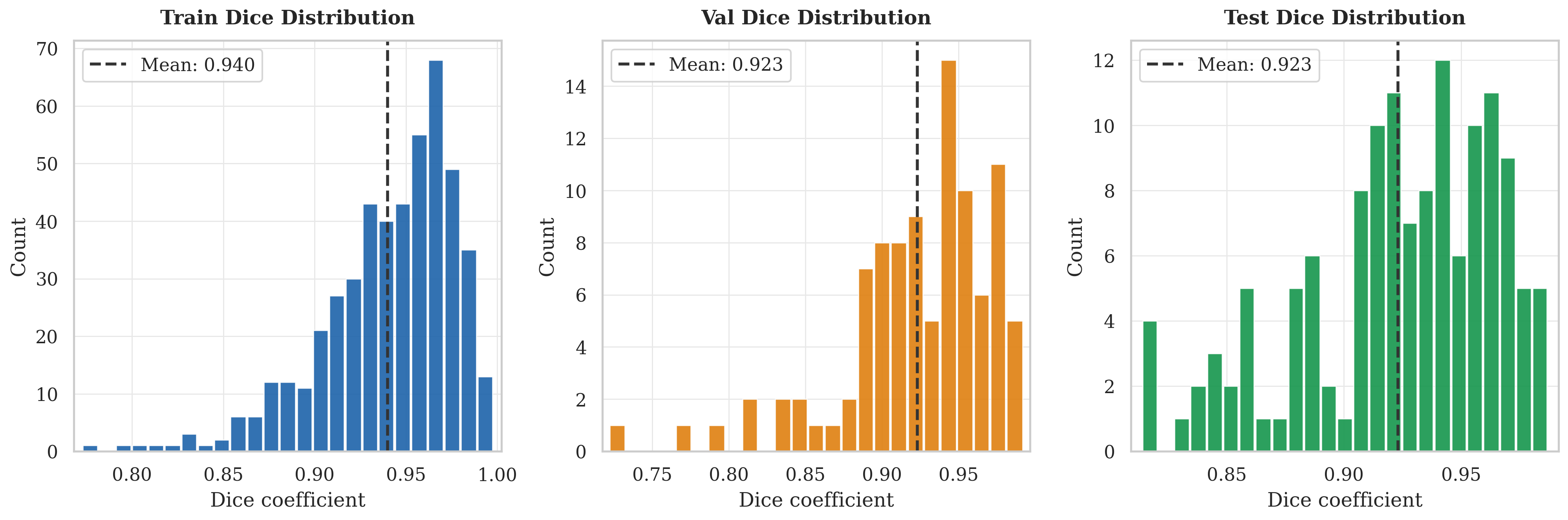}
\caption{Batch-level Dice distributions for training, validation, and test partitions. Dashed vertical lines mark the displayed means.}
\label{fig:dice-hist}
\end{figure*}

The per-batch Dice distributions in Figure~\ref{fig:dice-hist} have means of approximately 0.940 for training and 0.923 for both validation and test. Most batches occupy the upper portion of the range, but all splits have a lower tail. The tail is important: it indicates that a high mean does not eliminate difficult batches, which likely contain fragmented fields, strong built-environment confusers, or small foreground fractions. A scene-stratified error analysis is needed to attribute those low scores to specific landscape types rather than infer causes from the histogram alone.

\subsection{Training dynamics}

\begin{figure*}[t]
\centering
\includegraphics[width=\textwidth]{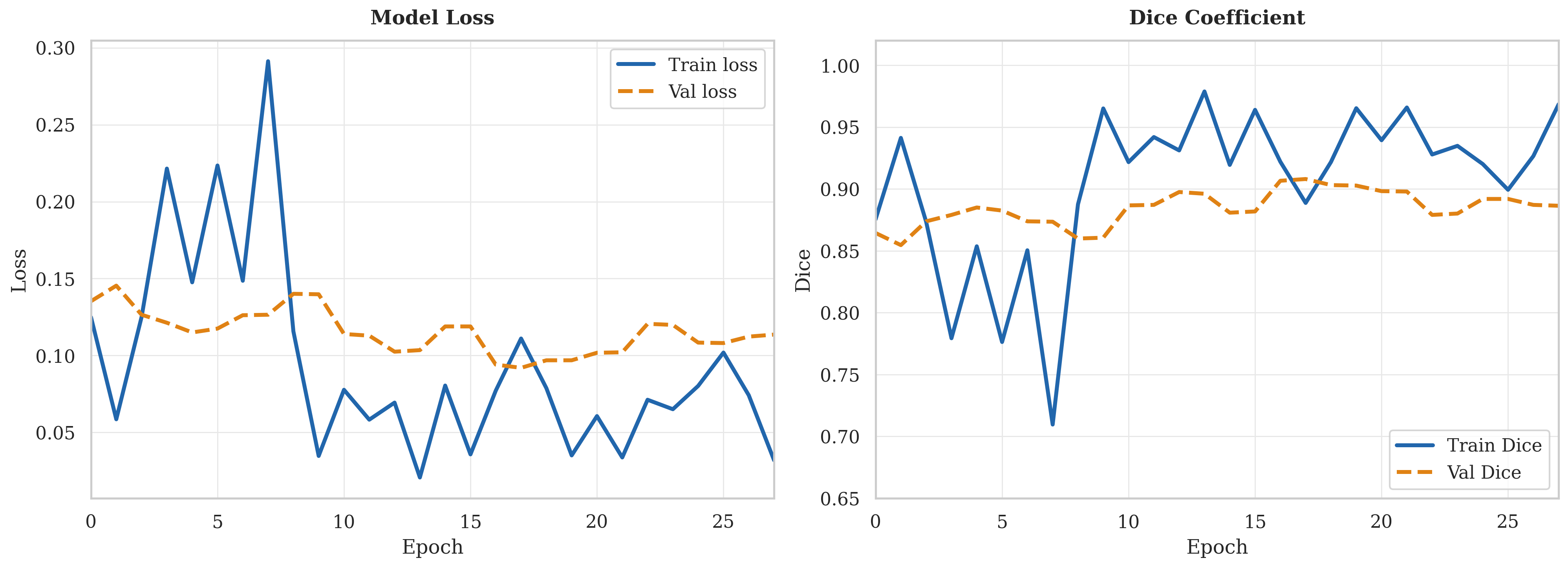}
\caption{Training history for model loss (left) and Dice coefficient (right) on the training and validation partitions.}
\label{fig:training-history}
\end{figure*}

Figure~\ref{fig:training-history} shows the recorded training and validation loss and Dice trajectories. Training values fluctuate more strongly than validation values during the early epochs, which is expected when mini-batches differ in farmland fraction, landscape complexity, and augmentation. After the initial adjustment period, validation Dice remains within a relatively narrow range and validation loss varies gradually. The plateau in the validation curves is consistent with checkpoint selection and early stopping: later epochs changed the training trajectory without producing a sustained improvement on the validation scenes.

\subsection{Selected SAM 3 refinements}

The frozen SAM 3 branch was examined on selected patches that exposed typical weaknesses of the ResUNet output. In an orchard-row example, the Dice coefficient increased from 0.858 for ResUNet alone to 0.955 after logical-OR fusion. In a fragmented-parcel example, Dice increased from 0.804 to 0.903. These cases represent gains of 0.097 and 0.099, respectively; across the selected difficult examples retained in the project record, the improvement range was 0.041--0.099.

\begin{table*}[t]
\centering
\caption{Selected difficult patches used to illustrate the behavior of SAM 3 refinement.}
\label{tab:sam3}
\begin{tabular*}{\textwidth}{@{\extracolsep{\fill}}lrrr@{}}
\toprule
\textbf{Landscape structure} & \textbf{ResUNet Dice} & \textbf{ResUNet + SAM 3 Dice} & \textbf{Change} \\
\midrule
Orchard rows        & 0.858 & 0.955 & +0.097 \\
Fragmented parcels  & 0.804 & 0.903 & +0.099 \\
\bottomrule
\end{tabular*}
\end{table*}

The examples show the mechanism intended by the hybrid design: SAM 3 adds coherent agricultural regions where the domain model under-segments a field or breaks a repeated crop pattern. Because the examples were selected to examine difficult structures, they are interpreted as case-based evidence rather than a dataset-wide effect estimate. Logical-OR fusion is most favorable when omission dominates; its behavior under false-positive SAM 3 proposals is considered in the Discussion.

\subsection{Stitched regional masks}

\begin{figure*}[t]
\centering
\includegraphics[width=\textwidth]{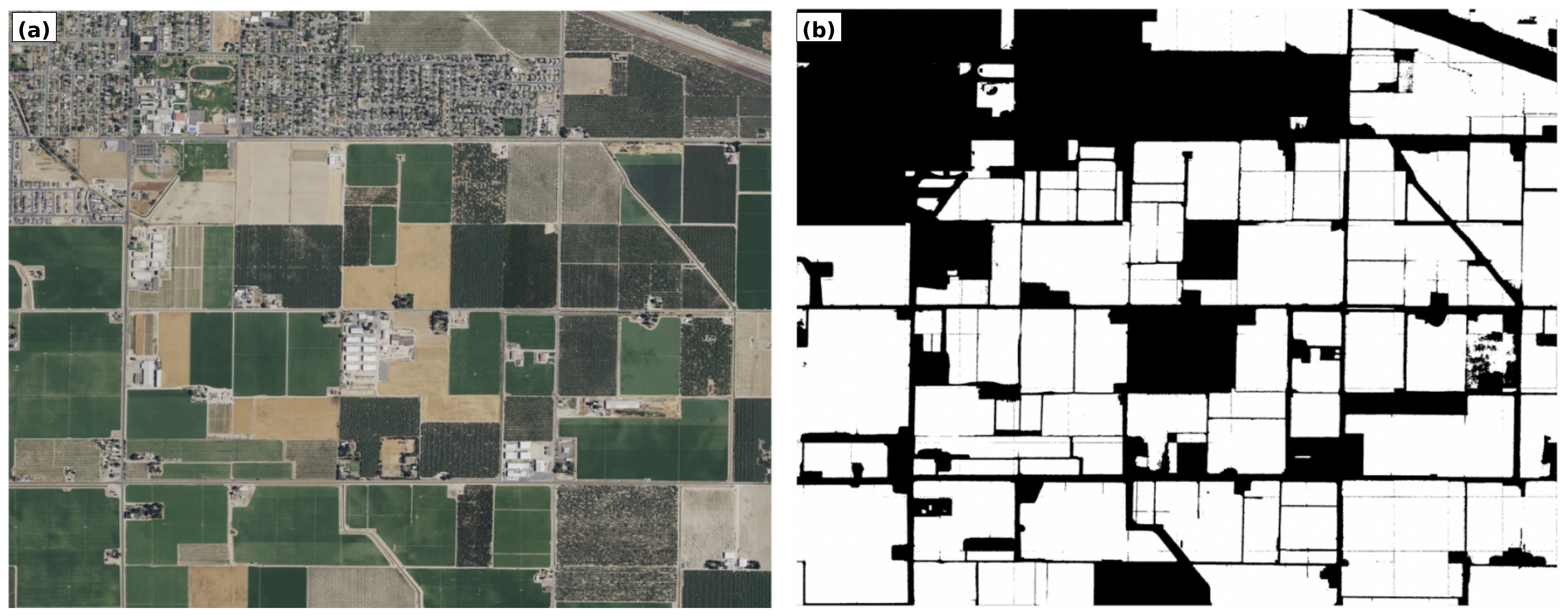}
\caption{Selected peri-urban mosaic. (a) NAIP RGB input. (b) Stitched fused mask, with farmland shown in white and background in black; Dice~=~0.898.}
\label{fig:periurban}
\end{figure*}

\begin{figure*}[t]
\centering
\includegraphics[width=\textwidth]{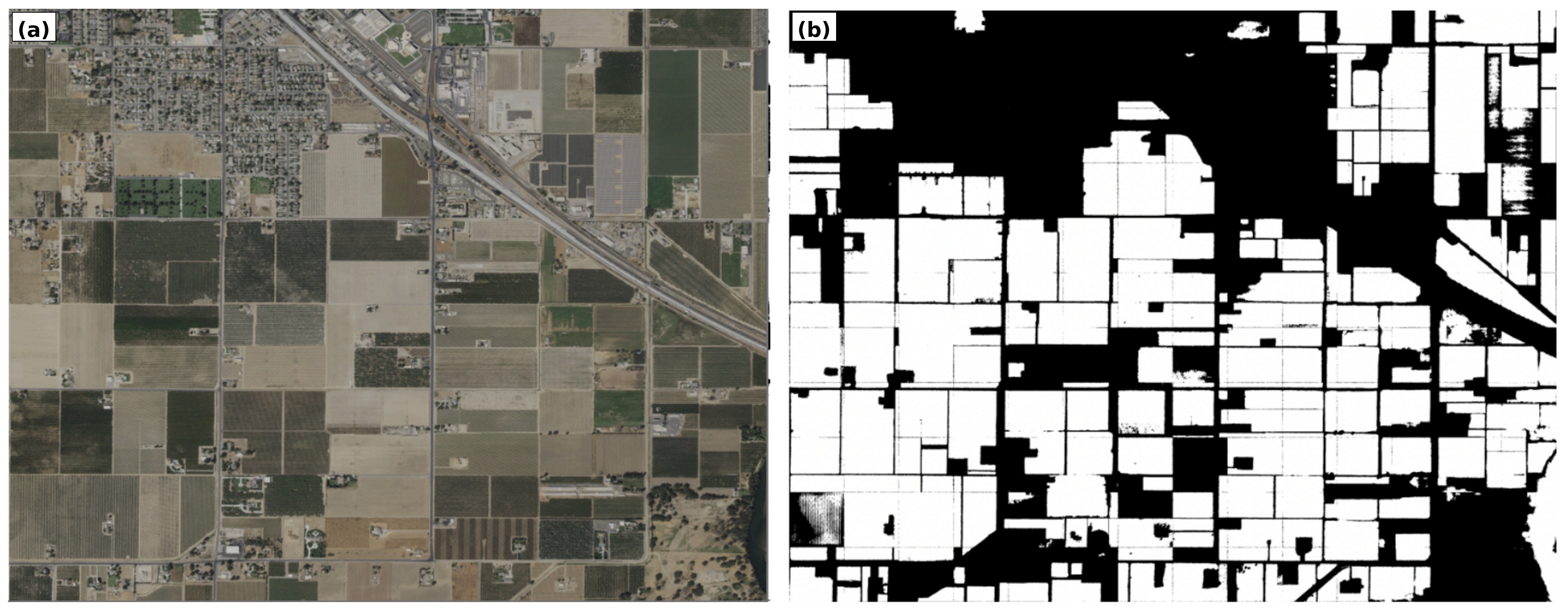}
\caption{Selected highway and urban-fringe scene. (a) NAIP RGB input. (b) Stitched fused farmland mask.}
\label{fig:highway}
\end{figure*}

\begin{figure*}[t]
\centering
\includegraphics[width=\textwidth]{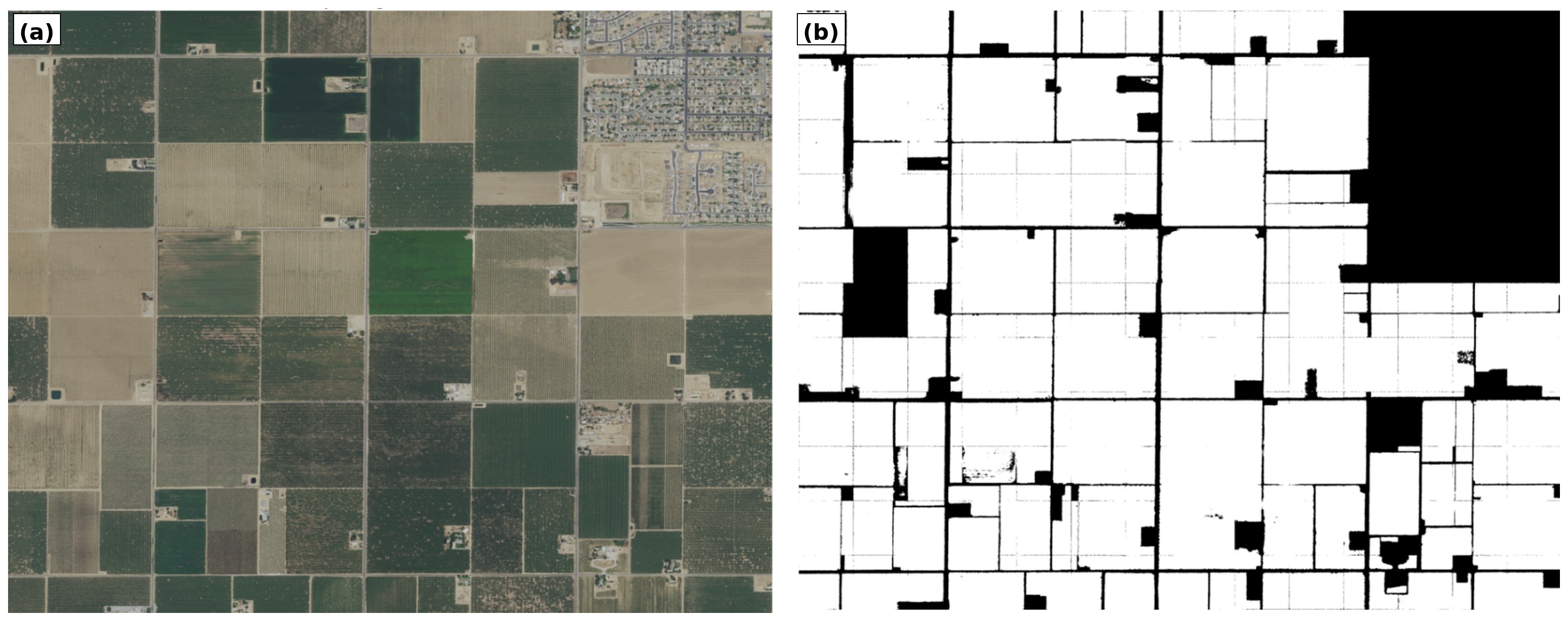}
\caption{Selected regular field-grid scene. (a) NAIP RGB input. (b) Stitched fused mask; Dice~=~0.919.}
\label{fig:fieldgrid}
\end{figure*}

Figures~\ref{fig:periurban}--\ref{fig:fieldgrid} show selected scene-scale predictions. The peri-urban mosaic in Figure~\ref{fig:periurban} includes dense residential development along the upper edge and a mixture of large and small agricultural blocks. The fused mask preserves many rectilinear field extents while excluding much of the developed area; the selected tile Dice is 0.898. Errors remain visible as small inclusions, omissions, and merged transitions, which are expected from a binary extent output.

Figure~\ref{fig:highway} presents a more difficult highway scene. The diagonal transportation corridor creates long, high-contrast edges and separates agricultural blocks from built-up land. The prediction largely excludes the main corridor and dense settlement while retaining fields on both sides. Some narrow linear features and developed parcels remain challenging, demonstrating why area-overlap metrics should be complemented by boundary and object-level assessment in future work.

The regular field grid in Figure~\ref{fig:fieldgrid} produces the highest selected tile score, 0.919. Large rectangular fields are reconstructed coherently across window boundaries, and the mask captures the dominant parcel geometry. The upper-right built-up area is excluded, although some field separations remain thin or merged. The contrast between Figures~\ref{fig:periurban} and \ref{fig:fieldgrid} is consistent with the expectation that regular, high-contrast geometry is easier than heterogeneous peri-urban structure.

\section{Discussion}\label{sec:discussion}

\subsection{A recall-oriented farmland mask}

The central quantitative result is not simply the test Dice of 0.9234; it is the combination of high overlap, very high recall, and lower precision. The model is tuned to avoid missing farmland. That behavior can be useful when the output is a screening layer for crop-area inventory or land-conversion monitoring, because omissions may remove fields from subsequent analysis. Yet over-prediction has costs. False positives can assign roads, construction sites, vacant land, lawns, or industrial surfaces to farmland and inflate area estimates. The appropriate operating point therefore depends on downstream use. A conservation screening tool may tolerate extra candidates that are reviewed later, whereas payment, insurance, or regulatory decisions require substantially stronger precision, calibration, and uncertainty reporting.

The loss formulation helps explain the trade-off. A 2.5-weighted Dice term responds strongly to large false-negative regions, whereas BCE retains local pixel supervision. This is a reasonable response to background-dominated patches, but it does not directly optimize boundary distance or object topology. Field-boundary literature has shown that multi-task predictions of extent, edges, and distance transforms can improve delineation and enable instance post-processing \citep{waldner2020deepedge,waldner2021decode}. Adding an explicit boundary head, signed-distance target, or topology-aware loss would be a more principled next step than relying on an extent mask alone.

\subsection{Contribution of text-prompted SAM 3}

SAM 3 introduces a qualitatively different prior from the ResUNet. The ResUNet learns the appearance of farmland from the annotated NAIP patches, whereas SAM 3 contributes open-vocabulary concept localization learned from a much broader visual corpus \citep{carion2025sam3}. The selected examples suggest that this prior is useful when repeated orchard texture, weak internal contrast, or fragmented geometry forms a coherent agricultural concept that the local classifier only partly recovers. This complementary behavior is consistent with remote-sensing studies that use SAM as a mask generator, promptable assistant, or auxiliary branch rather than as a complete replacement for domain supervision \citep{osco2023samrs,ao2024sempnet}.

The logical-OR rule favors completeness. Every positive ResUNet pixel is retained, and SAM 3 can only add candidate farmland. This makes the fusion transparent and explains the improvement observed in omission-dominated examples. The same asymmetry can also preserve a false-positive concept mask, so future refinements should consider confidence-weighted fusion, boundary gating, or selective SAM 3 invocation on low-confidence windows. Such alternatives would preserve the interpretability of the cascade while allowing evidence from the domain model to reject weak concept matches.

The additional foundation-model pass also has deployment implications. ResUNet can be applied rapidly over every patch, whereas SAM 3 is a larger model and is better suited to targeted refinement than indiscriminate use over very large archives. A practical regional workflow could therefore run ResUNet scene-wide and invoke SAM 3 only for patches with high uncertainty, broken field interiors, or unusually complex boundaries.

\subsection{Extent, visible boundaries, and parcel instances}

The distinction between semantic farmland extent and parcel delineation is central to interpreting the maps. A binary mask answers whether each pixel belongs to annotated farmland. It does not assign unique identities to adjacent fields, and it may merge parcels when the separating line is narrow or spectrally weak. Legal cadastral boundaries can also pass through visually uniform land and may be impossible to infer from imagery alone. The current product is therefore suitable as a farmland-area layer and as an input to later boundary or instance extraction, but it should not be presented as a cadastral substitute.

This distinction also affects evaluation. Dice and IoU measure area overlap and are insensitive to whether a predicted region contains the correct number of parcels. Boundary displacement metrics, object IoU, over-segmentation, under-segmentation, and topology are needed when the intended product is a parcel map \citep{wang2023survey}. Public datasets such as AI4Boundaries and Fields of the World provide useful structures for future benchmarking \citep{dandrimont2023ai4boundaries,kerner2025fields}, though their sensor resolution and geographic contexts differ from 1\,m NAIP.

\subsection{Landscape diversity and generalization}

The scene-selection strategy emphasized difficult backgrounds rather than maximizing the number of homogeneous crop patches. Highways, residential grids, industrial surfaces, school complexes, and irregular urban edges were intentionally represented because these features produce consequential commission errors in farmland maps. Scene-level partitioning further reduced the chance that neighboring content would appear in both training and evaluation. The close validation and test metrics therefore provide encouraging evidence that the model learned features that transferred across the sampled scenes rather than memorizing adjacent patches.

At the same time, NAIP appearance varies with acquisition year, season, local crop system, soil conditions, illumination, and state-level acquisition programs. The present scene collection was designed for landscape diversity, not for a formal continental probability sample. Broader transfer can be assessed by reserving entire regions or acquisition years for external evaluation and by reporting performance separately for open farmland, peri-urban scenes, semi-arid irrigation systems, and fragmented mosaics. This type of stratification is especially important because field-size and boundary visibility are known to influence segmentation difficulty across agricultural systems \citep{wang2022unlocking,tripathy2026zeroshot,robinson2026global}.

\subsection{Operational and policy relevance}

A defensible open farmland-extent layer can support several workflows. It can define the spatial support for crop-type probabilities, vegetation-index trajectories, yield estimates, and disease-risk signals. It can help local agencies screen where agricultural land is being replaced by roads, housing, or industrial development. It can provide candidate field geometry where proprietary layers are unavailable and enable researchers to aggregate NAIP or satellite observations to management units. The multiscale sensing literature makes clear why this step matters: leaf and canopy signals become actionable at regional scale only when they can be associated with locations that correspond to management decisions \citep{sishodia2020applications,weiss2020metareview}.

However, policy use raises a higher evidentiary bar than exploratory mapping. The model should not determine legal status, eligibility, or enforcement without human review and authoritative records. Temporal updating also requires change detection rather than independent segmentation alone. A field that is fallow, recently harvested, or temporarily disturbed may be visually ambiguous, and the binary label ``farmland'' embeds an annotation policy that must be documented. Releasing the labeling guide, ambiguous examples, and quality-control procedures is therefore as important as releasing model weights.

\subsection{Study scope and next steps}

Several aspects define the scope of the present results. First, the output is a binary semantic farmland mask. It represents visible farmland extent but does not assign unique parcel identities or reproduce legal cadastral boundaries. Boundary-aware heads, distance-transform targets, and polygon consolidation are natural extensions when an instance-level product is required.

Second, the SAM 3 evidence is deliberately case based. The selected improvements clarify how text-prompted concept masks can recover omissions, while a comprehensive paired comparison of ResUNet-only, SAM 3-only, and fused outputs would quantify how often the same operation improves or degrades precision, overlap, and boundary accuracy across the full test collection. Standard U-Net and DeepLabV3+ implementations would also provide useful architectural context under a common split and training budget.

Third, the model uses a single RGB acquisition. Near-infrared information and multi-date imagery may improve separation of cultivated land from bare ground, lawns, and non-crop vegetation, especially in dormant or recently harvested fields. Regional and acquisition-year hold-outs would further establish transfer beyond the current scene collection.

Finally, operational adoption requires measured throughput, memory use, and uncertainty. A selective cascade in which SAM 3 is triggered only on ambiguous patches may retain most of the observed benefit while controlling cost. Publishing the annotation policy, source-scene manifest, fixed data split, trained weights, and evaluation materials with a persistent identifier will make these extensions easier to compare and reproduce.

\section{Conclusions}\label{sec:conclusions}

This study developed a hybrid workflow for mapping farmland extent and visible boundaries from 1\,m NAIP RGB imagery. A residual U-Net trained on 5{,}698 scene-separated patches achieved a test Dice coefficient of 0.9234, IoU of 0.8605, accuracy of 0.8808, precision of 0.8766, and recall of 0.9794. The resulting mask is strongly recall oriented: it retains most annotated farmland while concentrating remaining errors at visually ambiguous infrastructure, exposed soil, and urban--agricultural transitions.

A frozen SAM 3 branch extended the domain model with the text prompt ``agricultural farmland field.'' Logical-OR fusion improved selected orchard-row and fragmented-parcel examples and supported coherent stitched masks in contrasting regional scenes, including a peri-urban mosaic with Dice 0.898 and a regular field grid with Dice 0.919. The central value of the design lies in its clear division of labor: ResUNet supplies efficient task-specific prediction, while SAM 3 contributes a broad semantic prior without additional farmland-specific training.

The workflow offers a practical foundation for crop-area estimation, agricultural monitoring, and screening of farmland conversion where current field layers are unavailable. Future work should quantify fusion performance across the full test collection, compare standard segmentation baselines, test external regions and acquisition years, incorporate NIR or multi-date observations, and develop boundary-aware vectorization for applications that require separate field instances. These extensions can build on the present semantic product without overstating its relationship to legal cadastral boundaries.

\section*{Acknowledgments}

The authors thank Anav Apparaju for his invaluable assistance with manual annotation of farmland polygons.

\section*{Funding}

The authors received no specific financial support for the research, authorship, or publication of this article.

\section*{Declaration of interest}

The authors report no competing interests.

\section*{Declaration of generative AI use}

During the preparation of this work, the authors used ChatGPT to improve grammatical accuracy, refine sentence structure, and enhance visualizations. All AI-generated revisions were thoroughly reviewed and edited by the authors to ensure relevance and accuracy.

\section*{Data availability statement}

NAIP source imagery is publicly available through the U.S. Department of Agriculture and U.S. Geological Survey distribution services (\url{https://doi.org/10.5066/F7QN651G}). The derived annotation masks, 256\,$\times$\,256 image patches, scene-level split manifest, and evaluation materials supporting this article are openly available in Zenodo at \url{https://doi.org/10.5281/zenodo.21519912}. The source code for ResUNet training, SAM 3 refinement, and tile stitching is available on GitHub at \url{https://github.com/MohammadrezaNarimaniUCDavis/NAIP_Farmland_ResSAM}.

\section*{Author contributions (CRediT)}

Mohammadreza Narimani: Conceptualization, Methodology, Software, Formal analysis, Investigation, Data curation, Visualization, Writing--original draft, Writing--review and editing, Project administration. Vikram Anand: Data curation, Investigation, Validation, Visualization, Writing--review and editing. Parastoo Farajpoor: Conceptualization, Methodology, Supervision, Validation, Writing--review and editing. All authors reviewed and approved the final manuscript and accept accountability for the work.

\section*{Ethics statement}

Not applicable. The study used aerial imagery and did not involve human participants, animals, or personally identifiable data.

\bibliographystyle{IEEEtranN}
\bibliography{references}

\end{document}